\documentclass[runningheads]{llncs}
\usepackage[T1]{fontenc}
\usepackage{amsmath}
\usepackage{amsfonts}
\usepackage{multirow}
\usepackage{booktabs}
\def\doi#1{\href{https://doi.org/\detokenize{#1}}{\url{https://doi.org/\detokenize{#1}}}}
\usepackage{graphicx}
\usepackage{color}
\usepackage{listings}

\begin{document}
\title{Calibrating Label Distribution for Class-Imbalanced Barely-Supervised Knee Segmentation}
\titlerunning{CLD-Semi}
\author{
Yiqun Lin\inst{1} \and
Huifeng Yao\inst{1} \and
Zezhong Li\inst{3} \and
Guoyan Zheng\inst{3}$^\star$ \and
Xiaomeng Li\inst{1, 2}\thanks{Corresponding Authors: {\tt\small eexmli@ust.hk}, {\tt\small guoyan.zheng@sjtu.edu.cn}}}
\authorrunning{Yiqun Lin \and
Huifeng Yao \and
Zezhong Li \and
Guoyan Zheng$^\star$ \and
Xiaomeng Li$^\star$}
\institute{
The Hong Kong University of Science and Technology \and
The Hong Kong University of Science and Technology Shenzhen Research Institute  \and
Shanghai Jiao Tong University
}
\maketitle

\newcommand{\yq}[1]{{\color[rgb]{0.1,0.1,0.9}{[#1]}}}

\newcommand{\xmli}[1]{{\color[rgb]{0.8,0.1,0.1}{[XM:#1]}}}
\newcommand{\nickname}{CLD}
\newcommand{\etal}{\textit{et al}.}

\begin{abstract}

Segmentation of 3D knee MR images is important for the assessment of osteoarthritis. Like other medical data, the volume-wise labeling of knee MR images is expertise-demanded and time-consuming; hence semi-supervised learning (SSL), particularly barely-supervised learning, is highly desirable for training with insufficient labeled data.
We observed that the class imbalance problem is severe in the knee MR images as the cartilages only occupy 6\% of foreground volumes, and the situation becomes worse without sufficient labeled data. 
To address the above problem, we present a novel framework for barely-supervised knee segmentation with noisy and imbalanced labels. Our framework leverages label distribution to encourage the network to put more effort into learning cartilage parts.
Specifically, we utilize 1.) label quantity distribution for modifying the objective loss function to a class-aware weighted form and 2.) label position distribution for constructing a cropping probability mask to crop more sub-volumes in cartilage areas from both labeled and unlabeled inputs. 
In addition, we design dual uncertainty-aware sampling supervision to enhance the supervision of low-confident categories for efficient unsupervised learning. 
Experiments show that our proposed framework brings significant improvements by incorporating the unlabeled data and alleviating the problem of class imbalance. More importantly, our method outperforms the state-of-the-art SSL methods, demonstrating the potential of our framework for the more challenging SSL setting. Our code is available at {\tt\small https://github.com/xmed-lab/CLD-Semi}.

\end{abstract}

\section{Introduction} \label{sec:intro}

\begin{figure}[t]
\centering 
\includegraphics[width=1.0\textwidth]{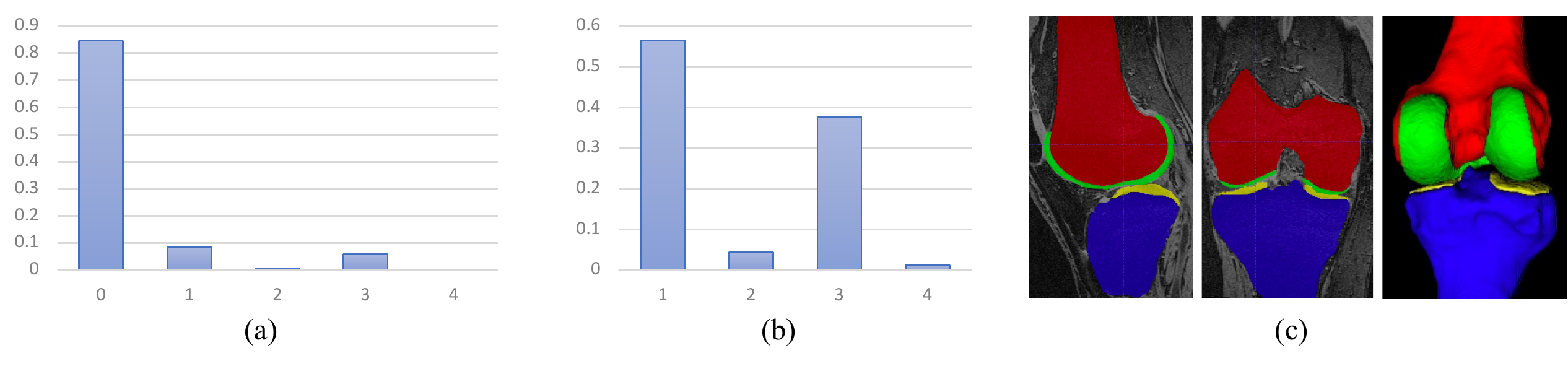}
\vspace{-0.9cm}
\caption{(a) Quantity distribution of all background (labeled as 0) and foreground categories (labeled as 1,2,3,4). (b) Quantity distribution of foreground categories including distal femur, femoral cartilage, tibia, and tibial cartilag, which are labeled as 1,2,3,4, respectively. (c) Visualization of segmentation ground truth (left and middle) and reconstructed mesh (right). Red, green, blue, yellow color refer to the above 4 categories, respectively.}
\label{fig:dist}
\end{figure}

The most common form of arthritis in the knee is osteoarthritis, a degenerative, ``wear-and-tear'' type of arthritis that occurs most often in people 50 years of age and older. Magnetic resonance imaging (MRI) is a widely used medical imaging technology~\cite{li20183d}. It is ideally suited for the assessment of osteoarthritis because it can clearly show soft-tissue contrast without ionizing radiation. For an objective and quantitative analysis, high-precision segmentation of cartilages from MR images is significant. 
With the development of deep learning technology, automatic knee segmentation has drawn more and more attention~\cite{ambellan2019automated,liu2018deep,prasoon2013deep}.
However, different from natural images, the segmentation of knee MR images suffers from a class imbalance problem. As shown in Figure~\ref{fig:dist}.a-b, the foreground volumes (cartilages and hard tissues) occupy 16\% of the entire image, and the cartilages only occupy 6\% of foreground volumes, which implies a severe class imbalance between foreground and background and between cartilage and hard tissue. 

Though deep learning methods can achieve better performance than morphological analysis, they require massive pixel-wise annotation to be trained in full supervision. In the medical field, sufficient labeled data is more difficult to obtain than natural images as manual annotation is expertise-demanded and time-consuming. Therefore, semi-supervised learning was introduced to solve this problem by utilizing only a small amount of labeled data and an arbitrary amount of unlabeled data for training. 
Recently, many semi-supervised learning (SSL) methods were proposed to solve insufficient labeled data problems on the natural images~\cite{chen2021semi,ding2021kfc,xxl,tarvainen2017mean,xie2020self} and medical images~\cite{huang20213d,li2018semi,li2020transformation,luo2021urpc,yao2022enhancing,uamt_yu}.
In particular, \cite{bai2017semi,zhou2019semi} are proposed to generate pseudo labels for unlabeled data with model parameter fixed for the next round training. \cite{tarvainen2017mean,uamt_yu} proposed to guide the model to be invariant to random noises in the input domain. \cite{chen2021semi,fang2020dmnet,luo2020semi,luo2021efficient} proposed to design several models or decoders and use consistency regularization for unsupervised learning. \cite{luo2021urpc,uamt_yu} leveraged the uncertainty information to enable the framework to gradually learn from meaningful and reliable targets.
Although appealing results have been achieved by these SSL methods, they cannot handle the class imbalance problem with barely labeled data. 
Recent work AEL (Adaptive Equalization Learning~\cite{hu2021semi}) proposed adaptive augmentation, re-weighting, and sampling strategies to solve the class imbalance for natural images in SSL. However, Table~\ref{tab:all} shows the improvement of AEL is limited since the proposed strategies are not suitable for medical data.

In this work, we aim to address the problem of class imbalance in semi-supervised knee segmentation with barely labeled data. 
We regard CPS (Cross Pseudo Supervision~\cite{chen2021semi}) as the baseline framework as it achieves the state-of-the-art performance on the SSL segmentation task for natural images. We further present a novel SSL framework named {\nickname} (\textbf{C}alibrating \textbf{L}abel \textbf{D}istribution) by leveraging the label distribution and uncertainty information to guide the model to put more effort into the learning of cartilage parts and enhance the learning of low-confident categories. Specifically, we firstly modify the objective loss function to a class-aware weighted form by utilizing the quantity distribution of labels. As shown in Figure~\ref{fig:dist}.c, the soft cartilages are much thinner than hard tissues and occupy fewer volumes along the z-axis (from up to down), resulting in the cartilages being less cropped in random cropping augmentation, which further exacerbates the class imbalance problem. Therefore, we propose probability-aware random cropping to crop more in cartilage areas of both labeled and unlabeled input images by incorporating the position distribution of labels. 
Furthermore, we observe that the output confidence of cartilage volumes is lower than hard tissues due to the class imbalance. Hence we design dual uncertainty-aware sampling supervision to enhance the supervision of low-confident categories (i.e., cartilages). Concretely, instead of using a constant sampling rate, We maintain an uncertainty bank for each of the two models to estimate the sampling rate for each category.

To summarize, the main contributions of this work include 1.) 
we are the first to address the class imbalance problem in barely-supervised knee segmentation;
2.) we propose a novel SSL framework {\nickname} for knee segmentation, consisting of class-aware weighted loss, probability-aware random cropping, and dual uncertainty-aware sampling supervision; 3.) we conduct extensive experiments and ablation studies to validate the effectiveness of the proposed methods on a clinical knee segmentation dataset.

\section{Method}

\begin{figure}[t]
\centering 
\includegraphics[width=1.0\textwidth]{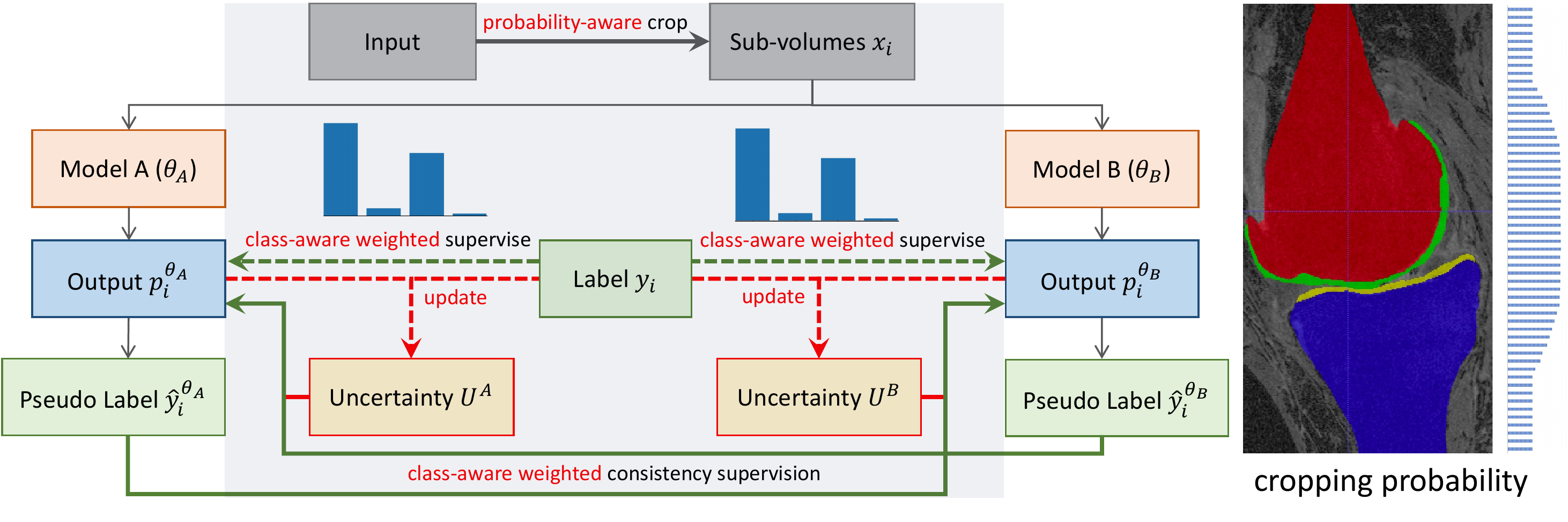}
\vspace{-0.9cm}
\caption{Overview of the proposed semi-supervised segmentation framework. We modify the original supervised/unsupervised loss to a weighted form by leveraging the quantity distribution of segmentation labels. We replace the random cropping with a probability-aware cropping strategy by incorporating the position distribution (right) of cartilages. In addition, we design dual uncertainty-aware sampling to enhance the supervision on low-confident categories for efficient unsupervised learning.}
\label{fig:framework}
\end{figure}

As illustrated in Figure~\ref{fig:framework}, our framework consists of two models with the same architecture but different initial parameters. We modify the objective loss function to a class-aware weighted form and replace the random cropping with probability-aware random cropping to address the problem of class imbalance. In addition, we design dual uncertainty-aware sampling supervision to enhance the supervision on low-confident categories by maintaining two uncertainty banks for two models.

\subsection{Cross Supervision for Semi-Supervised Segmentation}

In this work, we study the task of semi-supervised segmentation for knee MR imaging scans. We follow CPS~\cite{chen2021semi} to firstly initialize two models with the same architecture but different parameters $\theta_A$ and $\theta_B$, respectively. To formulate, let the labeled set be $\mathcal{D}_L=\big\{(x_i, y_i)\big\}^{N_L}_{i=1}$ with $N_L$ data and the unlabeled set be $\mathcal{D}_U=\{x_i\}^{N_U}_{i=1}$ with $N_U$ data, where $x_i \in \mathbb{R}^{H\times W\times D}$ is the input volume and $y_i \in \{0,1,2,3,4\}^{H\times W\times D}$ is the ground-truth annotation (4 foreground categories). Denote the output probability of the segmentation model as $p_i^\theta = f(x_i; \theta)$ and the prediction (pseudo label) as $\hat{y}_i^\theta= \text{argmax}(p_i^\theta)$, where $\theta$ indicates the model parameters. The goal of our semi-supervised segmentation framework is to minimize the following objective function:
\begin{equation}
    \mathcal{L} = \sum_{i=1}^{N_L} \Big[L_s(p_i^{\theta_A}, y_i) + L_s(p_i^{\theta_B}, y_i)\Big] + \lambda\sum_{i=1}^{N_L + N_U} \Big[\mathcal{L}_u(p_i^{\theta_A}, \hat{y}_i^{\theta_B}) + \mathcal{L}_u(p_i^{\theta_B}, \hat{y}_i^{\theta_A})\Big],
\end{equation}
where $\mathcal{L}_s$ is the supervised loss function to supervise the output of labeled data, and $\mathcal{L}_u$ is the unsupervised loss function to measure the prediction consistency of two models by taking the same input volume $x_i$. Note that both labeled and unlabeled data are used to compute the unsupervised loss. In addition, $\lambda$ is the weighting coefficient, ramping up from 0 to $\lambda_{\text{max}}$ for controlling the trade-off between the supervised loss and the unsupervised loss.

In practice, we employ V-Net~\cite{milletari2016v} as the backbone network and regard CPS~\cite{chen2021semi} as the SSL baseline framework. We follow \cite{uamt_yu} to remove the short residual connection in each convolution block. In the baseline, we use cross-entropy (CE) loss as the unsupervised loss, and a joint cross-entropy loss and soft dice loss as the supervised loss function, which are given as follows:
\begin{equation}
    \mathcal{L}_u(x, y) = \mathcal{L}_\text{CE}(x, y),\ \ \mathcal{L}_s(x, y) = \frac{1}{2} \Big[\mathcal{L}_\text{CE}(x, y) + \mathcal{L}_\text{Dice}(x, y)\Big].
\end{equation}
In addition, we empirically choose $\lambda_\text{max}$ as 0.1 and use the epoch-dependent  Gaussian ramp-up function
$
\lambda(t)=\lambda_{\max} * e^{-5\left(1 - \frac{t}{t_{\max }}\right)^{2}},
$
where $t$ is the current training epoch and $t_{\max }$ is the total number of training epochs.

\subsection{Calibrating Label Distribution (CLD)}

To solve the class imbalance problem in barely-supervised knee segmentation, we propose a novel framework {\nickname}, by leveraging the label distribution of soft cartilages and hard tissues for addressing the class imbalance problem.

\vspace{6pt} \noindent
\textbf{Class-aware weighted loss.} We firstly modify the supervised and unsupervised loss function to a weighted form by introducing class-aware weights. We utilize the category distribution of labeled data by counting the number of voxels for each category, denoted as $N_i, i=0, \dots, C$, where $C$ is the number of foreground categories, and $N_0$ indicates the number of background voxels. We construct the weighting coefficient $w_i$ for $i^\text{th}$ category as follows
\begin{equation}
    w_i = \Big(\frac{\max \{n_j\}_{j=0}^{C}}{n_i}\Big)^\alpha,\ \ n_i = \frac{N_i}{\sum_{j=0}^{C} N_j},\ \ i = 0, \dots, C.
\end{equation}
The exponential term $\alpha$ is empirically set to $\frac{1}{3}$ in the experiments. For cross-entropy loss calculation, the loss of each voxel will be multiplied by a weighting coefficient depending on using true label ($\mathcal{L}_s$) or pseudo label ($\mathcal{L}_u$). The soft dice loss will be calculated on the input image for each category separately and then multiplied by the weighting coefficient.

\vspace{6pt} \noindent
\textbf{Probability-aware random cropping.} As mentioned in Section~\ref{sec:intro}, the soft cartilages are thinner, and we propose probability-aware random cropping replacing random cropping to crop more sub-volumes in cartilage areas from both labeled and unlabeled inputs.
Since the distributions of foreground categories along x-axis and y-axis are quite similar, we only consider the cropping probabilities along z-axis (from up to down). Suppose that the total length is $D$ and the cropping size is $D'$ along z-axis. To formulate, for each labeled image $x_i$, we calculate a vector $v_i$ with the length of $D$, where the $j^\text{th}$ value of $v_i$ is 1 only when there are more than $k_1$ voxels labeled as soft cartilages in the cropping window centered at $j^\text{th}$ voxel along z-axis. Then we sum all $v_i$ to obtain $v = \sum_{i=1}^{N_L} v_i$ and increase the cropping probability by a factor of $\beta$ at $j^\text{th}$ position if $j^\text{th}$ value of $v$ is greater than $k_2$. In the experiments, we empirically choose both $k_1$ and $k_2$ as 1, and $\beta$ as 2.0.

\vspace{6pt} \noindent
\textbf{Dual uncertainty-aware sampling supervision.} To alleviate the uncertainty imbalance brought by class imbalance and limited labeled data, we adopt the sampling strategy to sample fewer voxels of low-uncertainty categories and more voxels of high-uncertainty categories for supervision. Instead of using a constant sampling rate, maintain an uncertainty bank for each category as $U \in \mathbb{R}^C$ for estimating the sampling rate on-the-fly. Assume that the output is $p \in \mathbb{R}^{C\times WHD}$ and the one-hot label is $y\in \mathbb{R}^{C\times WHD}$, then the uncertainty of $i^\text{th}$ category is given by
\begin{equation}
    u_i = 1 - \frac{\sum y_i \cdot p_i}{\sum y_i},
\end{equation}
where $p_i, y_i \in \mathbb{R}^{WHD}$ are $i^\text{th}$ values in the first dimension, indicating the prediction and label value for $i^\text{th}$ category. Due to the class-imbalance problem, the cropped sub-volumes sometimes cannot contain all categories. In practice, we accumulate the values of $\sum y_i \cdot p_i$ and $\sum y_i$ for $k_3$ times to obtain a more stable uncertainty estimation:
\begin{equation}
    u_i = 1 - \frac{\sum_{j=1}^{k_3}\sum y_i^j \cdot p_i^j}{\sum_{j=1}^{k_3}\sum y_i^j},
\end{equation}
where $p^j, y^j$ is the output and label of $j^\text{th}$ input sub-volume. In addition, the uncertainty values are initialized randomly and updated as an exponential moving average (EMA) with a monmentum $\gamma$, i.e., $u_i^t = \gamma u_i^{t-1} + (1-\gamma) u_i^{t'}$. Note that the uncertainty values are only measured from the output of labeled data, and we maintain two uncertainty banks for two models respectively. Then we define the sampling rate for each category as $s_i = \left(\frac{u_i}{\max_i u_i}\right)^{1/2}$. Taking the supervision on model A as an example. Let pseudo labels from model B be $\hat{y}^{\theta_B}$, and the uncertainty bank and sampling rates of model A be $U^A = \{u_1^A, \dots, u_C^A\}$ and $S^A = \{s_1^A, \dots, s_C^A\}$, respectively. For those voxels predicted as $i^\text{th}$ category in $\hat{y}^{\theta_B}$, we randomly sample a subset of voxels with the sampling rate $s_i^A$ and construct the binary sampling mask as $m_i^A\in\{0,1\}^{WHD}$. Compute all $m_i^A$ and denote the union sampling mask as $m^A = \bigcup m_i^A$. Therefore, only the voxels with the value of 1 in $m^A$ (i.e., sampled voxels) will contribute to the unsupervised loss. $k_3$ and $\gamma$ are empirically set to 8 and 0.999 in the experiments.

\section{Experiments}

We conducted comprehensive experiments to validate the effectiveness of our proposed methods on a collected knee segmentation dataset. In addition, we conduct extensive ablation experiments to analyze the working mechanism of different proposed modules in the framework.

\vspace{6pt} \noindent
\textbf{Dataset and pre-processing.} 
We collected a knee segmentation dataset with 512 MR imaging scans, containing 412 for training, 50 for validation, and 50 for testing. 
The size of each imaging scan is 384$\times$384$\times$160.
All the image data are publicly available from Osteoarthritis Initiative (OAI\footnote{https://oai.nih.gov}). Ground-truth segmentation of the data were done by orthepaedic surgeons from local institution.
There are 4 foreground categories, including distal femur (DF), femoral cartilage (FC), tibia (Ti), and tibial cartilage (TC), which have extremely imbalanced distribution. Some example slices and the category quantity distribution are shown in Figure~\ref{fig:dist}.
We follow the previous work \cite{uamt_yu} to normalized the input scans as zero mean and unit variance before being fed into the network.

\vspace{6pt} \noindent
\textbf{Implementation.} We implement the proposed framework with PyTorch, using a single NVIDIA RTX 3090 GPU. The network parameters are optimized with SGD with a momentum of 0.9 and an initial learning rate of 0.01. The learning rate is divided by $0.001^{1/300} \approx 0.9772$ per epoch. Totally 300 epochs are trained as the network has well converged. The batch size is 4, consisting of 2 labeled data and 2 unlabeled data.
We choose 160$\times$160$\times$48 as the cropping size of sub-volumes in the training and testing. In the inference (testing) stage, final segmentation results are obtained using a sliding window strategy~\cite{uamt_yu} with a stride size of 64$\times$64$\times$16, where the outputs of overlapped volumes are averaged over all windows' outputs. Standard data augmentation techniques~\cite{yu2017automatic,uamt_yu} are used one-the-fly to avoid overfitting, including randomly flipping, and rotating with 90, 180 and 270 degrees along the axial plane.

\vspace{6pt} \noindent
\textbf{Evaluation metrics and results.} We evaluate the prediction of the network with two metrics, including Dice and the average surface distance (ASD). We use 4 scans (1\%) as labeled data and the remaining 408 scans as unlabeled data. In Table~\ref{tab:all}, we present the segmentation performance of different methods on the testing set. The first two rows show the results of V-Net~\cite{milletari2016v} trained with the full training set and with only 1\% labeled data, revealing that the lack of efficient labels makes the class imbalance problem worse, and brings a dramatic performance drop. By utilizing the unlabeled data, our proposed SSL framework significantly improves the performance in all categories. To validate our network backbone design (V-Net~\cite{milletari2016v}), we also conduct the experiments with nnUNet~\cite{isensee2021nnu}, and the final testing Dice score is 90.8\% (Avg.) with the full labeled set (412 data) for training, which shows that we can regard the revised V-Net~\cite{milletari2016v} as a standard backbone model.

\begin{figure}[t]
\centering 
\includegraphics[width=1.0\textwidth]{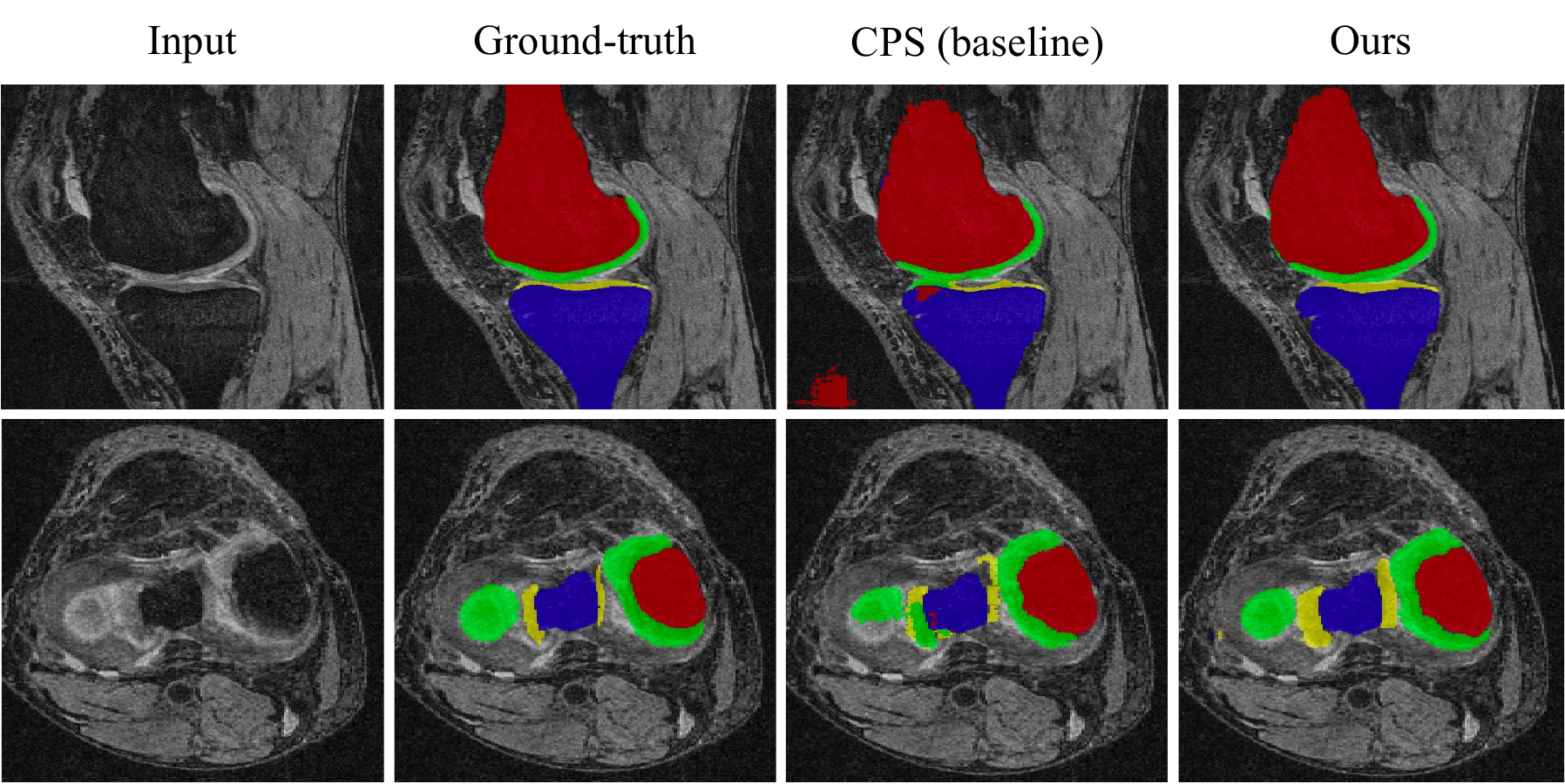}
\vspace{-0.8cm}
\caption{Comparison of segmentation results with CPS~\cite{chen2021semi}. As the cartilages (colored in {\color[rgb]{0.2, 0.9, 0.2}{green}} and {\color[rgb]{0.9, 0.9, 0.1}{yellow}}) are much thinner and tightly connected to hard tissues, CPS does not perform well in the junction areas of cartilages and hard tissues, while ours can.}
\label{fig:res}
\end{figure}

Furthermore, we implemented several state-of-the-art SSL segmentation methods for comparison, including UA-MT~\cite{uamt_yu}, URPC~\cite{luo2021urpc}, and CPS~\cite{chen2021semi} in Table~\ref{tab:all}. Although utilizing the unlabeled data, the Dice scores of two cartilages are still much worse than hard tissues. In addition, we adopted the learning strategies in AEL~\cite{hu2021semi} to CPS, but the improvement is still limited.
Compared with the baseline model (CPS), our proposed learning strategies further improve the performance by 3.8\% Dice on average. The results also show that the proposed framework alleviates the class imbalance problem and improves the performance by 2.6\% and 10.9\% Dice for two cartilages, respectively. Visual results in Figure~\ref{fig:res} show our method can perform better in the junction areas of cartilages and hard tissues.


\begin{table}[t]
\caption{Comparison between our method with previous methods. DF: distal femur, FC: femoral cartilage, Ti: tibia, and TC: tibial cartilage.} \label{tab:all}
\vspace{-6pt}
\renewcommand\tabcolsep{1pt}
\centering
\begin{tabular}{l|cc|c|cccc}
\toprule[1.5pt]
\multirow{2}{*}{Method} & \multicolumn{2}{c|}{\# scans used} & \multicolumn{5}{c}{Dice [\%]$\uparrow$ / ASD [voxel]$\downarrow$} \\ \cline{2-8} 
 & labeled & unlabeled & Avg. & DF & FC & Ti & TC \\ \hline
\multirow{2}{*}{V-Net~\cite{milletari2016v}} & 412 & 0 & 90.5/3.4 & 97.2/7.4 & 86.4/1.2 & 97.3/3.7 & 81.1/1.3 \\
 & 4 & 0 & 25.9/- & 68.3/35.1 & 69.9/13.4 & 0.0/- & 0.0/- \\ \hline
UA-MT~\cite{uamt_yu} & 4 & 408 & 32.0/- & 67.7/51.5 & 60.1/24.2 & 0.0/- & 0.0/- \\
URPC~\cite{luo2021urpc} & 4 & 408 & 76.6/26.2 & 88.7/26.9 & 74.4/5.2 & 82.6/46.7 & 60.5/25.9 \\
CPS~\cite{chen2021semi} & 4 & 408 & 83.4/16.3 & 93.1/17.2 & 81.1/2.4 & 91.5/24.5 & 67.7/20.9 \\
CPS+AEL~\cite{chen2021semi,hu2021semi} & 4 & 408 & 83.6/15.1 & 93.2/16.8 & 81.3/2.7 & 90.8/27.3 & 69.2/13.5 \\ \hline
{\nickname} (ours) & 4 & 408 & \textbf{87.2}/\textbf{8.8} & \textbf{93.8}/\textbf{14.9} & \textbf{83.7}/\textbf{1.1} & \textbf{92.8}/\textbf{17.9} & \textbf{78.6}/\textbf{1.2} \\
\bottomrule[1.5pt]
\end{tabular}
\end{table}

\vspace{6pt} \noindent
\textbf{Analysis of our methods.} To validate the effectiveness of the proposed learning strategies, including class-aware weighted loss (WL), dual uncertainty-aware sampling supervision (DUS), and probability-aware random cropping (PRC), we conduct ablative experiments, as shown in Table~\ref{tab:abs}. We can see that WL improves the Dice scores of two cartilages 1.1\% and 9.3\% but brings a 2.3\% performance drop for hard tissues,
which means WL can improve the learning of cartilages, but in turn negatively affect the learning of hard tissues.
DUS maintains the improvements on cartilages and alleviates the performance drop of hard tissues. PRC can further boost the performance of both soft cartilages and hard tissues.

\begin{table}[t]
\caption{Ablative study on different proposed modules. WL: class-aware weighted loss function, DUS: dual uncertainty-aware sampling supervision, and PRC: probability-aware random cropping.} \label{tab:abs}
\vspace{-6pt}
\renewcommand\tabcolsep{6.15pt}
\centering
\begin{tabular}{ccc|ccccc}
\toprule[1.5pt]
\multirow{2}{*}{WL} & \multirow{2}{*}{DUS} & \multirow{2}{*}{PRC} & \multicolumn{5}{c}{Dice [\%]$\uparrow$ / ASD [voxel]$\downarrow$} \\ \cline{4-8} 
 &  &  & \multicolumn{1}{c|}{Avg.} & DF & FC & Ti & TC \\ \hline
 &  &  & \multicolumn{1}{c|}{83.4/16.3} & 93.1/17.2 & 81.1/2.4 & 91.5/24.5 & 67.7/20.9 \\
$\surd$ &  &  & \multicolumn{1}{c|}{84.8/14.4} & 90.8/23.9 & 82.2/1.5 & 89.2/31.0 & 77.0/1.1 \\
$\surd$ & $\surd$ &  & \multicolumn{1}{c|}{86.1/9.7} & 92.5/\textbf{11.4} & 82.5/1.3 & 91.0/24.9 & 77.1/\textbf{1.0} \\
$\surd$ & $\surd$ & $\surd$ & \multicolumn{1}{c|}{\textbf{87.2}/\textbf{8.8}} & \textbf{93.8}/14.9 & \textbf{83.7}/\textbf{1.1} & \textbf{92.8}/\textbf{17.9} & \textbf{78.6}/1.2 \\
\bottomrule[1.5pt]
\end{tabular}
\end{table}
\section{Conclusion}


In this work, we propose a novel semi-supervised segmentation framework {\nickname} by introducing class-aware weighted loss (WL), probability-aware random cropping (PRC), and dual uncertainty-aware sampling supervision (DUS) to enhance the learning and supervision in the areas of cartilages in knee MR images. 
Among them, WL and DUS are general solutions for solving the class imbalance problem in semi-supervised segmentation tasks. PRC is the specific design for the knee dataset, where the cartilages are extremely thin and have a smaller cropping probability along the z-axis than hard tissues.
Extensive experiments show that the proposed framework brings significant improvements over the baseline and outperforms previous SSL methods by a considerable margin.

\vspace{6pt}
\noindent
\textbf{Acknowledgement.} This work was supported by a grant from HKUST-Shanghai Jiao Tong University (SJTU) Joint Research Collaboration Fund (SJTU21EG05), a grant from HKUST-BICI Exploratory Fund (HCIC-004), and a grant from Shenzhen Municipal Central Government Guides Local Science and Technology Development Special Funded Projects (2021Szvup139).

\nocite{*}
\bibliographystyle{splncs04}
\bibliography{miccai.bib}

\end{document}